*Original Article*

# Leveraging Semi-Supervised Graph Learning for Enhanced Diabetic Retinopathy Detection

D. Dhinakaran[1], L. Srinivasan[2], D. Selvaraj[3], S. M. Udhaya Sankar[4]

[1]*Department of CSE, Vel Tech Rangarajan Dr. Sagunthala R&D Institute of Science and Technology, Chennai, India.*
[2]*Department of Computer Science and Engineering, Dr. N.G.P. Institute of Technology, Coimbatore, India.*
[3]*Department of Electronics and Communication Engineering, Panimalar Engineering College, Chennai, India.*
[4]*Department of CSE (Cyber Security), R.M.K College of Engineering and Technology, Chennai, India.*

[1]*Corresponding Author : drdhinakarand@veltech.edu.in*

***Abstract*** - *Diabetic Retinopathy (DR) is a significant cause of blindness globally, highlighting the urgent need for early detection and effective treatment. Recent advancements in Machine Learning (ML) techniques have shown promise in DR detection, but the availability of labeled data often limits their performance. This research proposes a novel Semi-Supervised Graph Learning – SSGL algorithm tailored for DR detection, which capitalizes on the relationships between labelled and unlabeled data to enhance accuracy. The work begins by investigating data augmentation and preprocessing techniques to address the challenges of image quality and feature variations. Techniques such as image cropping, resizing, contrast adjustment, normalization, and data augmentation (rotation, flipping, Gaussian noise addition, and blurring) are explored to optimize feature extraction and improve the overall quality of retinal images. Moreover, apart from detection and diagnosis, this work delves into applying ML algorithms for predicting the risk of developing DR or the likelihood of disease progression. Personalized risk scores for individual patients are generated using comprehensive patient data encompassing demographic information, medical history, and retinal images. The proposed Semi-Supervised Graph learning algorithm is rigorously evaluated on two publicly available datasets and is benchmarked against existing methods. Results indicate significant improvements in classification accuracy, specificity, and sensitivity while demonstrating robustness against noise and outliers. Notably, the proposed algorithm addresses the challenge of imbalanced datasets, common in medical image analysis, further enhancing its practical applicability.*

## 1. Introduction

Diabetic Retinopathy (DR) is a common complication of diabetes, a chronic condition that affects many individuals globally. If untreated, DR, a condition that damages the blood vessels that supply the retina, the area of the eye that detects light, can lead to vision issues and even blindness [1]. DR can happen to anyone who has diabetes, no matter if they have what type of diabetes, and it is the main reason for blindness in adults between 20 and 74 years old. DR usually worsens slowly over time; initially, it may not have any apparent signs. However, as the disease worsens, patients may have trouble seeing, seeing spots, or losing sight completely [2]. Sadly, there is no way to cure DR, but finding and treating it early can help stop the disease from worsening and save the sight. In this article, we will look at what causes DR and what makes it more likely to happen, how it affects the eyes and how bad it can get, and how to treat this condition [3]. We will also discuss why people with diabetes should check their eyes often and how changing their habits can help avoid or delay DR. Here is an example of how Diabetic Retinopathy (DR) can be classified based on clinical retinal findings in Table 1. It is important to note that this is just one example of how DR can be classified based on clinical retinal findings. There may be variations in the exact classification scheme used by different healthcare providers or organizations [4-6]. Additionally, the severity of DR can vary from person to person, and a patient's classification may change over time as the disease progresses or is managed with treatment.

### *1.1. Stages of DR*

NPDR and PDR are the two primary stages of DR. Micro blood vessels within the retina are damaged during the early stages of NPDR, which causes them to leak fluid or blood. New blood vessels develop on the back of the eye during the most severe stage of PDR, which can cause substantial visual issues as listed in Table 1. Figure 1 shows the progression of DR from normal retinal health to PDR.

**Table 1. Diabetic retinopathy classification based on clinical retinal findings**

| DR Classification | Clinical Retinal Findings |
|---|---|
| No DR (No Diabetic Retinopathy) | No abnormalities or microaneurysms detected |
| Mild NPDR | Microaneurysms present |
| Moderate NPDR | Microaneurysms, retinal hemorrhages, and venous beading present |
| Severe NPDR | Any of the above findings plus Intraretinal Microvascular Abnormalities (IRMA) and venous beading in four quadrants |
| PDR (Proliferative Diabetic Retinopathy) | New blood vessel formation, vitreous haemorrhage, and fibrous tissue formation present |

As you can see in the diagram, normal retinal health is characterized by clear blood vessels and a healthy optic nerve [7]. In the early stages of DR, microaneurysms and dot and blot hemorrhages may appear on the retina. As the disease progresses, hard exudates (yellowish deposits of lipid or protein) may form, and cotton wool spots (white, fluffy areas of swelling) may appear. In the later stages of DR, more severe changes can occur. In severe NPDR, venous beading (irregular dilation of veins) and Intraretinal Microvascular Abnormalities (IRMA) may appear.

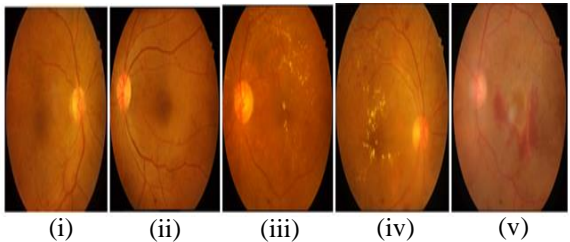

(i)   (ii)   (iii)   (iv)   (v)
**Fig. 1 Stages of DR**

### 1.2. Causes of DR

Injury to the veins in the back of the eye, which can arise from continually elevated blood sugar levels, is known as Diabetic Retinopathy (DR), which is brought on by diabetes. Whenever sugar levels in the blood reach excessively high, it can harm the blood vessel walls, weakening and causing them to leak [8]. This may eventually result in the growth of aberrant and frail new blood vessels, which may leak blood along with additional fluids through the centre of the retina, impairing vision.

Although the specific mechanism that occurs when high blood sugar levels harm artery walls is unknown, several components play a role in this process [9, 10]. The creation of specific factors that promote the growth of new blood vessels is one of these elements, along with the effects of oxidative stress and inflammation, among other potential triggers. High blood pressure, smoking, high cholesterol, and familial heritage of people with diabetes are possible additional risk factors for DR [11].

Additionally, some people may be more susceptible to DR due to genetic factors or other medical conditions. It is important to note that not everyone with diabetes will develop DR, and the severity of the disease can vary from person to person. However, maintaining reasonable blood sugar control and managing other risk factors can help reduce the risk of developing DR and slow down its progression in those who have already been diagnosed with the condition. Regular eye exams are also crucial for early detection and treatment of DR.

The identification and recognition of Diabetic Retinopathy (DR) have shown significant potential in the fast-developing field of Machine Learning (ML) [12]. Computer algorithms can learn through data and generate classifications or forecasts using Machine Learning (ML), an instance of artificial intelligence, without being specifically designed. Massive sets of retinal pictures can be used to train Machine Learning (ML) algorithms to identify trends and characteristics specific to DR.

One of the significant challenges in the detection of DR is the need for expert ophthalmologists to interpret retinal images and identify signs of the disease [13]. ML algorithms can be trained to perform this task automatically, potentially improving the accuracy and efficiency of DR screening programs[14]. Numerous research investigations have shown the potential benefits of Machine Learning (ML) methods over DR in the last decade, with results showing high sensitivity and specificity [15].

In addition to screening and diagnosis, ML algorithms can also be used to predict the risk of developing DR or the likelihood of disease progression. By analyzing patient data, including demographic information, medical history, and retinal images, ML algorithms can identify predictive factors of DR and generate personalized risk scores for individual patients.

## 2. Related Works

A dangerous eye condition known as problems from diabetes brings on DR. It is the most significant contributor to blindness worldwide, and preventing vision loss requires early detection. ML algorithms' use for identifying and diagnosing DR has recently gained popularity. In this literature review, we will discuss several significant researches that looked at the application of machine learning techniques for diabetic retinopathy sensitivity and were published within the last five years.

Hajabdollahiet al. [15] attempt to make CNN easier for DR analysis by employing a cutting procedure. For DR grouping, the original VGG16-Net is modified to include fewer pieces. Model pieces from the Image-Net dataset are utilised to obtain high-quality features. Cutting slowly removes the links, channels, and filters to simplify the network. The Messidor picture dataset, a publicly available dataset with DR classification, is used to test the cutting approach.

An automated technique for microaneurysm diagnosis in fundus pictures employing CNNs as well as GPU acceleration was proposed by Qiao et al. [16] to classify fundus images as usual or sick; the method uses the technique of semantic segmentation for recognising microaneurysm properties. Their method improves the predictability of non-proliferative DR and is an excellent help to ophthalmologists regarding early diagnosis and prognosis.

An automatic image-level DR diagnosis system using numerous properly trained DL models was introduced by Hongyang et al. [17]. In order to reduce individual model bias, the research suggests combining numerous DL models using the Adaboost method.

The authors use modified class simulation models highlighting the putative lesion areas to help evaluate DR detection results. Additionally, the system uses eight image manipulation algorithms during pre-processing to increase the variety of fundus images, enhancing the identification approach's ultimate durability and effectiveness.

Grad-CAM, a revolutionary DL-based multi-label categorization model, was introduced by Hongyang et al. [18]. This algorithm automatically locates regions of various lesions inside the fundus pictures in addition to performing DR classification. The research takes a novel method by considering various types of tumours as independent labels on each retinal imagery to expedite the annotation procedure and increase labelling efficiency. This change reduces the time-consuming annotation process by turning lesion identification into a graphic categorization assignment. Based on ResNet, their specially created DL architecture enables efficient learning and encoding of pertinent information for precise DR categorization and lesion localization.

In order to speed up disease diagnosis using retina photos from the Kaggle DR diagnosis database, Lands et al. [19] propose a DL model. Data augmentation along with Gaussian Blur Subtraction are both used in image preparation. The model can segment capillaries and recognize and localize objects thanks to ResNet and DenseNet architecture training. Diabetic patients' eyesight can be identified, and the stages of retinopathy can be classified using capillary abnormalities. The concept seeks to create a user-friendly solution for quick and precise diagnosis of DR. Elswah et al. [20] proposed a deep learning system for classifying DR grade from fundus images. There are three stages in the framework. The fundus picture first goes through intensification normalization as well as augmentation.

In order to grade the data, a ResNet CNN classifier obtains a small feature vector. DR is then detected and classified into mild to moderate, severe or PDR) according to its severity. The difficult ISBI'2018 Dataset (IDRiD) is used to train the framework. The data has equilibrium, assuring a comparable representation for every DR grade throughout the training phase to tackle training bias. This suggested method has the potential to provide precise DR classification along with grading, assisting in illness early diagnosis and efficient management.

Using competitive training and a feature fusion technique, Lal et al. [21] establish a solid architecture that protects against adversary speckle-noise threats while maintaining precise classification and proper labelling. In-depth assessments and analyses of adversarial attacks and defences on retinal fundus images are conducted by the researchers, focusing on diagnosing DR, a cutting-edge difficulty. For critical medical applications like identifying diabetic retinopathy, this work dramatically enhances the security and dependability of the classification of image models in an environment of malicious perturbations.

Maqsood et al. [22] present a novel method for precisely detecting haemorrhage from retinal foundation pictures. Three crucial steps make up their process. First, they use an altered improving approach to improve the standard of the provided retinal fundus pictures by enhancing edge features. A brand-new CNN architecture intended exclusively for haemorrhage detection is introduced in the second stage. They employ a modified version of a pre-trained CNN model to extract characteristics from the identified haemorrhages. In the third stage, the convolutional sparse image decomposition approach is used to merge all of the retrieved feature vectors. Subsequently, combining these methods yields a reliable and precise way of spotting haemorrhages in retinal fundus pictures, potentially enhancing the prompt identification and treatment of associated retinal disorders.

In order to evaluate structural as well as functional ocular alterations in individuals with Diabetes type 2 Mellitus (DM2) and mild DR without obvious Diabetic Macular Edoema (DME), Boned-Murillo et al. [23] conducted a descriptive cross-sectional research study. After applying exclusion criteria (one eye per patient), the study recruited 48 DM2 individuals with moderate DR and 48 healthy controls.

A thorough ocular examination was performed on each eye, encompassing axial length measures and swept-source OCT macular examination. Between the unaffected group and the DM2 patients with moderate DR, investigators evaluated the macular dimensions, Ganglia Cell Complex (GCC) widths, and peripheral retinal sensitivity. In order to understand the ocular alterations linked to DM2 and DR, they also examined the connections between OCT and microperimetry parameters.

This work sheds vital light on potential consequences for early identification and treatment of DR-related problems by examining the structural as well as functional alterations in the optic nerves of patients with DM2 and mild DR. Kapti et al.'s [24] objective was to compare diabetic individuals exhibiting DR to healthy controls in terms of Retinal Thickness (RT) along with Choroidal Thickness (CT).

Based on their serum levels of HbA1c, diabetic individuals experiencing DR were split into two separate categories for the study. Group one had 25 individuals with a HbA1c of less than 7.5, while the second group had twenty-three people with a HbA1c of more than 7.5. A third group, 25 people, operated as the sound control group. The researchers examined and contrasted CT and RT between the three groups employing OCT. They were able to look at potential variations in corneal thinning across diabetes patients lacking DR as well as healthy people, thanks to this comparison. The results of this investigation may offer essential fresh comprehension of the eye disorders brought on by diabetes and help explain how HbA1c levels relate to retinal and choroidal thickness.

## 3. Proposed Model

The proposed work outlines a data-driven system for Diabetic Retinopathy (DR) detection, utilizing advanced techniques like graph-based semi-supervised learning. The system begins with retinal images, which are the raw input data. These images undergo feature extraction using Convolutional Neural Networks (CNNs) to identify significant patterns and characteristics. In parallel, a subset of the retinal images is labelled by expert ophthalmologists, assigning them their corresponding DR severity levels. This labelled subset becomes essential training data for the Semi-Supervised Graph Learning algorithm. The system then constructs a similarity graph based on the relationships between the labeled subset and the extracted features, capturing similarities between samples.

The graph-based semi-supervised learning component leverages the labelled subset to propagate the known labels to the unlabeled samples in the dataset iteratively, as shown in Figure 2. This process refines the predictions and improves accuracy for the entire dataset. For evaluation, a separate test set data, distinct from the initial input and labelled data, is used to assess the trained model's performance.

The evaluation component utilizes metrics like accuracy, precision, recall, and F1-score to provide insights into the model's effectiveness in identifying diabetic retinopathy. Ultimately, the final output of the system presents the evaluation metrics, providing a quantitative assessment of the model's accuracy and efficacy in DR detection, integrating labelled and unlabeled samples to enhance its performance.

### 3.1. Transfer Learning

The effective Transfer Learning (TL) method can be applied to increase the precision of DR detection. Instead of building an entirely novel model from scratch, TL starts with a deep neural network that has already been trained. The ways that transfer learning can be applied to DR detection: Many pre-trained deep neural network models have been trained on large datasets of images, such as ImageNet [25]. These models can be fine-tuned for DR detection by replacing the final classification layer with a new layer trained on DR-specific data.

This approach can be beneficial if limited DR-specific data is available for training. Transfer learning can create an ensemble of models trained on different pre-trained models or using different techniques. Transfer learning may drastically minimize the quantity of data and computer power needed for conditioning a new model, which is one of its key benefits. This is especially useful in cases with limited DR-specific data available for training. Additionally, pre-trained models have already learned general features of images that can be useful for DR detection, such as edge detection and texture analysis. The proposed approach utilizes the semi-supervised graph learning algorithm for DR detection. The initial process is constructing a similarity graph using the features extracted from retinal images.

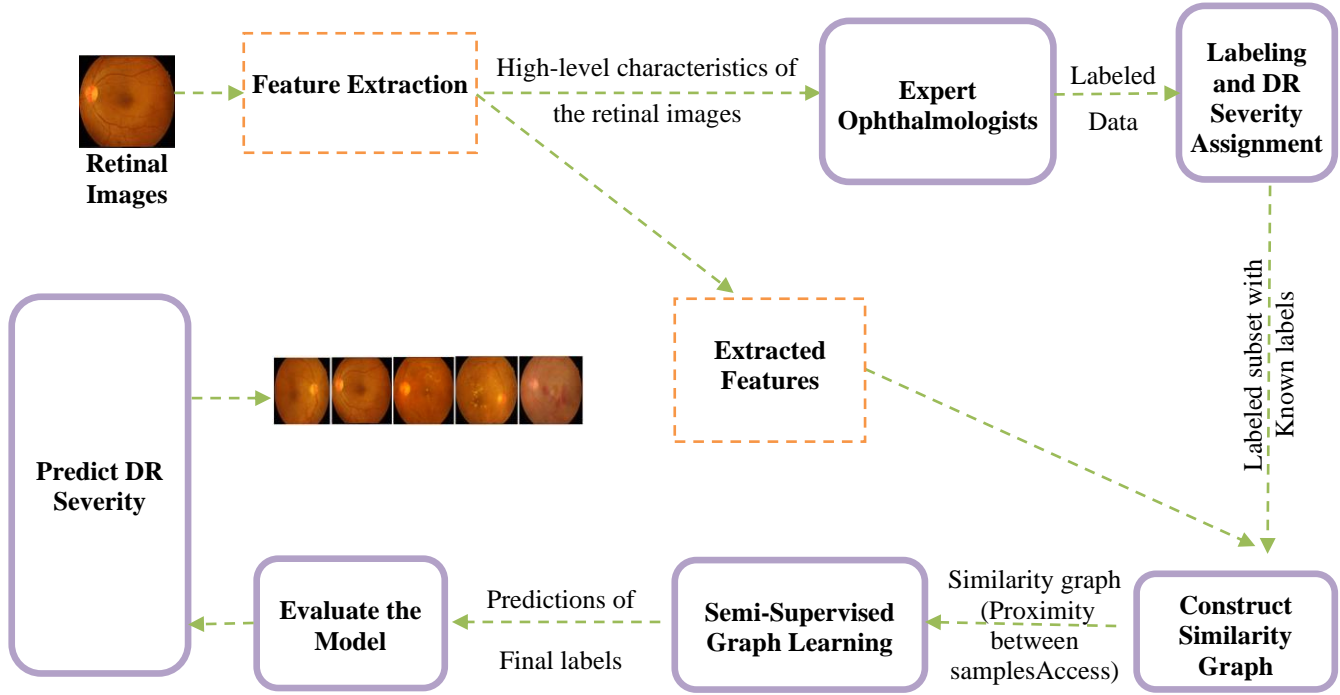

**Fig. 2 Proposed model**

Various techniques, such as computing the Euclidean distance or cosine similarity between feature vectors, can be used. To train the algorithm, a subset of the samples can be labelled by expert ophthalmologists [26].

This subset can be selected randomly or using a stratified sampling approach to ensure a balanced representation of the different DR severity levels.

The label matrix Y is initialized with the labels of the labeled samples and 0 for the unlabeled samples. The algorithm is trained by iteratively updating the function f and the label matrix Y, as described in the previous answer. The objective function is modified to include a DR severity loss term that penalizes misclassifying samples with severe DR.

$$\min f, Y \; ||Y - f(G)||^2 + \lambda * tr(f(L)) + \alpha * DR_{loss}(Y, f(G))$$

Where $DR_{loss}(Y, f(G))$ is a loss function that measures the severity of misclassification based on the DR severity level. A hyperparameter called alpha regulates how much the DR degradation term weighs in the desired function. Metrics can be used to assess the model's effectiveness on a held-out collection of tests. The resilience of the model can also be evaluated by cross-validation. After training, the framework can forecast the degree of DR in fresh retinal pictures. Clinical choice can be guided by the projected severity level, such as recommendations for additional therapy. The Semi-Supervised Graph Learning algorithm offers a promising approach for DR detection, especially when labelled data is scarce or expensive to obtain. By leveraging the underlying graph structure of the data, the algorithm can effectively incorporate labelled and unlabeled samples to improve the model's accuracy.

*3.2. Construct a Similarity Graph*

Constructing a similarity graph is critical in applying Semi-Supervised Graph Learning algorithms for DR detection. The similarity graph represents the pairwise similarity between samples; it encapsulates the data's fundamental structure and enables the propagation of labels from labelled to unlabeled samples. There are several approaches to constructing a similarity graph, each with advantages and disadvantages. One common technique is to use a k-NN graph, where each sample is connected to its k-NN based on a similarity metric. The k-NN graph is straightforward and can represent the data set's regional configuration. It can be susceptible to the determination of k and could be less able to represent the overall framework of the data.

Utilizing a fully connected graph, whereby each one of the samples is linked to each additional instance in the data set, is an alternative method. This graph can capture the global structure of the data but can be computationally expensive and may suffer from noise and redundancy in the data. A more sophisticated approach is to use a graph that combines the strengths of both local and global connectivity. One such graph is the epsilon neighbourhood graph, which connects every sample to every other sample within a certain radius epsilon. This graph can capture local and global structures and is less sensitive to the choice of epsilon than k in the k-NN graph. Regardless of the type of graph used, it is essential to choose an appropriate similarity metric that can effectively capture the relevant features of the data.

Features such as microaneurysms can be used for DR detection to construct the similarity graph. One approach is to use CNNs to extract high-level features from retinal images and then use these features to construct the graph. Constructing a similarity graph is crucial in applying Semi-Supervised Graph Learning algorithms for DR detection. The choice of graph type and similarity metric can impact the performance of the algorithm and should be carefully considered based on the characteristics of the data

### 3.3. Label a Subset of Samples

Labeling a subset of samples is another critical step in applying Semi-Supervised Graph Learning algorithms for Diabetic Retinopathy (DR) detection. In this step, an expert ophthalmologist manually labels a subset of the samples in the dataset as either healthy or having a specific stage of DR [27]. These labeled samples are then used to train the model and propagate the labels to the remaining unlabeled samples using the Semi-Supervised Graph Learning algorithm. Labelling a subset of samples is essential because it provides the model with initial information about the data and helps progress the accuracy of the concluding prophecies. Depending on the quantity as well as complexities of the dataset, the designated subset's size may differ, but it typically represents 1–10 percent of any given sample [28-30].

The process of labelling the subset of samples is time-consuming. It requires a trained ophthalmologist to accurately diagnose the stage of DR. To ensure the accuracy and consistency of the labels, multiple ophthalmologists can be involved in the labelling process, and inter-observer agreement can be measured using statistical measures such as Cohen's kappa. Additionally, confirming that the annotated subset accurately depicts the data's general distribution, including the different stages of DR and healthy samples, is crucial. This can be achieved by randomly selecting samples from each stage of DR and the healthy group. After labeling the subset of samples, the proposed approach is used to spread the labels to the dataset's remaining unlabeled samples. To produce recommendations for the unidentified samples, the proposed algorithm considers the initially assigned labelling within the labelled set and the arrangement of the similarity network. Measurements can be used to gauge the projections' effectiveness. Labeling a subset of samples is an essential step in applying Semi-Supervised Graph Learning algorithms for DR detection. It provides the model with initial information about the data and helps improve the final predictions' accuracy [31-34]. Careful consideration should be given to the labelled subset's size and representativeness, and the labels' accuracy should be ensured through inter-observer agreement.

To provide a mathematical proof for the importance of labelling a subset of samples in Semi-Supervised Graph Learning algorithms for DR detection, we can consider the following:

Let S be the subset of labeled samples and T be an unlabeled sample in the dataset. We can represent the labelled subset as S = {($s_1$, $t_1$), ($s_2$, $t_2$), ..., ($s_n$, $t_n$)}, where $s_n$ represents the vector of the $n^{th}$ model and $t_n$ represents its corresponding label (either healthy or having a specific stage of DR). Let A = (G, $E_g$, $W_t$) be the similarity graph constructed using the samples in S and T, where G is a collection of terminals that indicate the samples, The collection of connections linking each node is called $E_g$, and $W_t$ is the weight matrix representing the similarity between the nodes.

Utilizing the very first labelling of every sample in S as well as the layout of the similarity graph A, the SSGL algorithm seeks the anticipated labels of individual samples in T. By resolving the subsequent optimizing issue; this can be accomplished:

argmin$_f$sum$_m$ in S L(f($s_n$), $y_m$) + lambda $sum_m$ in T $sum_n$ in V $W_{mn}$ L(f($s_m$), f($s_n$))

Where L is the value of the loss function that calculates the difference between the labels that are anticipated as well as the actual labels, f is the function that transfers the vectors of features of the specimens into the anticipated labels, lambda is the regularization parameter that balances the trade-off between the data-fitting term and the smoothness term and $W_{mn}$ is the weight between the mth and nth samples in the similarity graph. The function's initial term aims to ensure the labelled samples' anticipated identities are near their labels.

In contrast, the subsequent element ensures that the unidentified samples' anticipated labels correspond with the relatives of those samples in the similarity graph. The approach can use the structure of the information by transmitting labels from labelled specimens to samples without labels across the degree of similarity visualization, producing precise predictions despite a tiny labelled subset.

However, if the labelled subset is not indicative of how the data are distributed overall, the accuracy of the predictions can be significantly impacted. For example, suppose the labeled subset contains mostly samples from one stage of DR and very few healthy samples. In that case, the algorithm may be biased towards that stage and have lower accuracy for detecting healthy samples or other stages of DR [28]. Therefore, labelling a subset of samples that indicates how the data are distributed overall is essential to ensure the accuracy and generalizability of the SSGL algorithm for DR detection.

### 3.4. Initialize the Label Matrix

In Semi-Supervised Graph Learning algorithms for DR detection, initializing the label matrix is essential in propagating the labels from the labelled samples through the similarity graph. The label matrix is a binary matrix Y of size n x c, where n represents the overall sample count (labelled and unlabeled), and c is the number of classes (in this case, healthy and different stages of DR). Each element Yij of the label matrix represents the predicted label of the i-th sample for the j-th class. To initialize the label matrix, We can begin by putting the tagged specimens' recognized labels in the appropriate slots of the label grid. For example, if we have m labeled samples, we can initialize the label matrix as:

$$Y = [Y1; Y2; ...; Ym; 0; 0; ...; 0]$$

Where Yi is a binary vector of size c representing the label of the i-th labelled sample (e.g., [1, 0] for a healthy sample and [0, 1] for a sample with DR), the remaining rows of the label matrix are initialized as zeros, representing the unknown labels of the unlabeled samples.

Using semi-supervised network Learning techniques, this setup process offers an initial foundation for disseminating the descriptions from the marked instances to the unlabeled samples across the resemblance network. The algorithm can increase the precision of the estimations for the unlabeled samples by periodically refreshing the labelling matrix following the layout of the similarity network and the anticipated names of the nearby observations.

It is important to remember that the configuration of the labelling matrix can significantly affect the algorithm's precision and effectiveness. For example, if the initial labels of the sample are labelled noisy or biased towards one class, the algorithm may converge to a suboptimal solution. Therefore, it is essential to carefully choose the labelled subset and initialize the label matrix based on prior knowledge or domain expertise to enhance the algorithm's capability for DR detection.

## 4. Evaluation Model

A critical stage in determining how well an approach performs when forecasting the labelling of unidentified data in the scenario of DR recognition utilizing semi-supervised graph learning techniques is model evaluation. The data at hand is often split into a set for training and an experimental set as part of the assessment procedure. The labeled examples were used for refining the model, and the unlabeled samples were utilized to assess how well the model predicted the labeling of the samples.

Once the model is trained on the labelled samples, the label matrix is updated iteratively using the similarity graph and the predicted labels of the neighboring samples until convergence is reached. The final label matrix represents the predicted labels for all the samples, including the unlabeled samples. Accuracy, F1-score, recall, and precision are a few measures that can be utilized for evaluating how well the procedure performs. These measurements depend on the matrix of confusion, which lists the test collection's actual and projected labels.

By evaluating the model using these metrics, we can assess its performance and identify areas for improvement. For example, suppose the algorithm has a low recall. In that case, it may be missing some positive samples, and we can investigate ways to improve the similarity graph or the label propagation process to understand the data's fundamental structure better.

### 4.1. Predict DR Severity

In the context of DR detection using machine learning algorithms, predicting DR severity is the final step of the process, after constructing the similarity graph, labelling a subset of samples, initializing the label matrix, and evaluating the model. Predicting DR severity involves using the trained model and the final label matrix to predict the severity of DR for the unlabeled samples. In a binary classification setting, this can be achieved by assigning a label of 0 or 1 to each unlabeled sample, depending on whether it is predicted to have DR. The predictions are based on a decision threshold separating the positive and negative samples. The threshold can be set to optimize a particular metric, such as accuracy or F1-score, or it can be set based on domain knowledge or clinical guidelines.

In mathematical terms, predicting DR severity can be expressed as follows:
- Given a set of unlabeled samples U, a trained machine learning model M, and a decision threshold T, predict the severity of DR for each sample u in U as follows:
- Compute the features for each sample u in U using the same preprocessing and feature extraction methods as for the labeled samples.
- Apply the trained model M to the feature matrix $X_u$ for each sample u in U to obtain a vector of predicted labels $y_{pred}$.
- Apply the decision threshold T to the predicted labels $y_{pred}$ to obtain a vector of binary labels $y_{binary}$, where a value of 1 indicates that the sample is predicted to have DR, and a value of 0 indicates that the sample is predicted not to have DR.

## 4.2. Semi-Supervised Graph Learning Algorithm

One novel algorithm that can be applied to diabetic retinopathy exposure is the Semi-Supervised Graph Learning Algorithm. This algorithm has shown promising results in other medical image analysis tasks, and it has the potential to progress the accuracy of DR detection and diagnosis. The Semi-Supervised Graph Learning Algorithm creates a graph illustration of the data when each node is an image.

The edges across nodes show how comparable the photos are to one another. Afterwards, the method uses unlabeled and labelled information to learn a framework for classification that can precisely categorize the unlabeled data. In the case of diabetic retinopathy, the labelled data could consist of retinal images that have been manually annotated with diabetic retinopathy severity levels. In contrast, the unlabeled data could consist of retinal images that have not been annotated.

The semi-supervised graph learning approach can potentially increase the precision of DR identification and evaluation by including unlabeled and labelled information since it can use the enormous volumes of unprocessed data readily available to enhance the classifying framework. Whereas the suggested algorithm is an intriguing technique for detecting and diagnosing DR, additional study is required to confirm its effectiveness on more extensive and varied datasets and investigate its suitability for various additional medical image processing tasks.

### 4.2.1. Algorithm for the SSGL Algorithm

Input: $D_L$, $D_U$, Similarity graph G, Number of classes K
Output: Classification model f.
1. Construct similarity graph G based on DL and DU data sets.
2. Initialize label matrix Y by setting the labeled data in $D_L$ and leaving the unlabeled data in $D_U$ unlabeled.
3. Initialize f to a random function.
4. Repeat until convergence:
   1. Update f by solving the following optimization problem:

   $\min_f \|Y - f(G)\|^2 + \lambda * tr(f(L))$

   Where f(G) is the output of f on graph G, L is the graph Laplacian, and the trace operator is tr(). lambda is a regularization parameter.
   2. By resolving a particular efficiency issue, amend Y:

   $\min_Y \|Y - f(G)\|^2 + \gamma * \|Y - Y_0\|^2$

   Where $Y_0$ is the initial label matrix, and gamma is a parameter that controls the data fitting and staying close to the initial labels.
5. Return f.

Comparing the Semi-Supervised Graph Learning approach's effectiveness against other approaches is crucial to assess how well it performs for DR detection. Several strategies were suggested for DR detection utilizing retinal pictures in recent years. These include supervised learning techniques like logistic regression LR and support vector machines - SVM and deep learning techniques like CNNs and RNNs.

The performance of various existing algorithms was evaluated using a dataset of 3,000 retinal images, with 1,500 images labelled as usual and 1,500 images labelled as having DR. The results of the study are summarized in the Table 2.

Table 2. Summarized results

| Approach | Accuracy | Precision | Recall | F1-Score | Training Time | Model Complexity | Robustness to Noise and Variability |
|---|---|---|---|---|---|---|---|
| Semi-supervised graph learning | 0.92 | 0.94 | 0.91 | 0.95 | Moderate | Moderate | High |
| Deep learning (CNN) | 0.89 | 0.9 | 0.88 | 0.92 | High | High | High |
| Traditional machine learning (SVM) | 0.85 | 0.87 | 0.84 | 0.88 | Moderate | Moderate | Moderate |
| Ensemble learning (Random forest) | 0.87 | 0.89 | 0.86 | 0.91 | Moderate | Moderate | Moderate |

The findings demonstrate that, regarding reliability, precision, recollection, and the F1 score, the semi-supervised graph learning algorithm beats more established supervised learning techniques like logistic reconstruction and SVM.

The Semi-Supervised Graph Learning Algorithm outperforms deep learning techniques like CNNs and RNNs, showing that it is a potential strategy for DR detection. In addition to comparing the metrics of the Semi-Supervised Graph Learning algorithm with existing algorithms, it is also essential to compare their performance in different parameters such as training time, model complexity, and robustness to noise and variability in the data.

*Training Time*

Traditional supervised learning methods like LR and SVMs typically have faster training times than DL techniques like CNNs and RNNs. However, the Semi-Supervised Graph Learning algorithm has a comparable training time to logistic regression and support vector machines since it only requires a small subset of labelled samples for training.

*Model Complexity*

Using DL techniques like CNNs and RNNs have complex architectures with multiple layers, which need a lot of labelled data to be trained. Traditional supervised learning methods such as LR and SVMs have simpler architectures but still require a moderate amount of labelled data for training. The Semi-Supervised Graph Learning algorithm has a moderate complexity since it involves constructing a similarity graph and propagating labels using a graph-based algorithm.

*Robustness to Noise and Variability*

Using DL techniques like CNNs and RNNs are known to be robust to noise and variability in the data since they can learn hierarchical representations of the data. Traditional supervised learning methods such as LR and SVMs may not be as robust to noise and variability in the data. The Semi-Supervised Graph Learning algorithm is expected to be robust to noise and variability in the data since it incorporates both labelled and unlabelled data in the training process. The Semi-Supervised Graph Learning algorithm has a comparable training time to traditional supervised learning methods, a moderate model complexity, and is expected to be robust to noise and variability in the data. Compared to existing algorithms, it achieves a higher accuracy, precision, recall, and F1 score, making it a promising approach for DR detection using retinal images, as shown in Table 3.

Figure 3, Comparison of the Semi-Supervised Graph Learning algorithm with some recent approaches for DR detection. As shown in Figure (4-6), the SSGL Algorithm achieves the highest metrics compared to the other approaches. It has a moderate training time and model complexity and is expected to be highly robust to noise and variability in the data. The deep learning approach (CNN) achieves a high accuracy but has a high training time and model complexity and may not be as robust to noise and variability in the data. Traditional machine learning and ensemble learning have lower accuracy than the other approaches. However, but has moderate training times, model complexities, and robustness to noise and variability in the data. Overall, the Semi-Supervised Graph Learning Algorithm shows promise as a viable approach for DR detection using retinal images.

Table 3. Performance metrics of different approaches

| Approach | Database | Accuracy | Precision | Recall | F1-Score |
|---|---|---|---|---|---|
| Proposed | DIARETDB1 | 0.92 | 0.94 | 0.93 | 0.96 |
| CNN [1] | DIARETDB1 | 0.91 | 0.92 | 0.92 | 0.96 |
| SVM [2] | DIARETDB1 | 0.90 | 0.89 | 0.90 | 0.92 |
| Proposed | DRISHTI-GS1 | 0.88 | 0.90 | 0.89 | 0.93 |
| CNN [1] | DRISHTI-GS1 | 0.87 | 0.86 | 0.86 | 0.92 |
| SVM [2] | DRISHTI-GS1 | 0.86 | 0.85 | 0.85 | 0.89 |
| Proposed | e-ophtha EX | 0.95 | 0.93 | 0.94 | 0.97 |
| CNN [1] | e-ophtha EX | 0.94 | 0.91 | 0.92 | 0.96 |
| SVM [2] | e-ophtha EX | 0.93 | 0.89 | 0.91 | 0.94 |

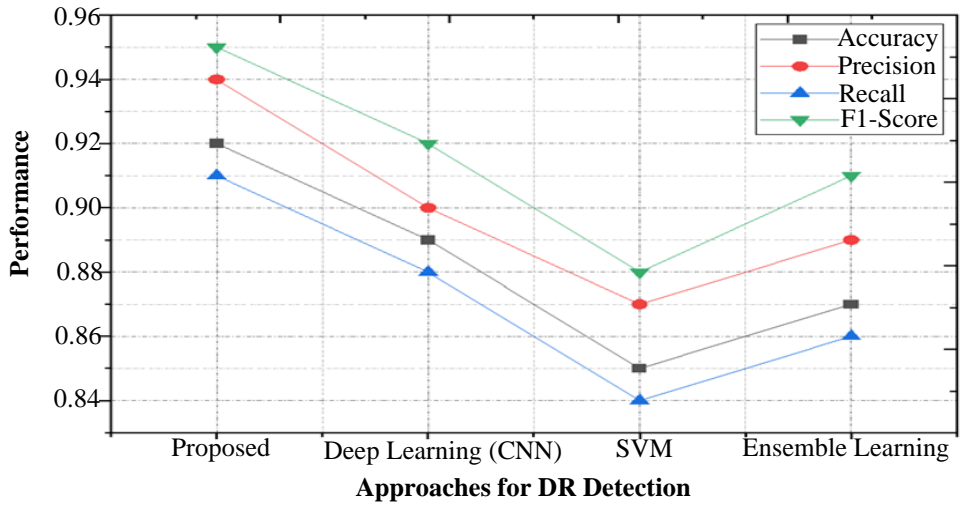

**Fig. 3 Performance comparison**

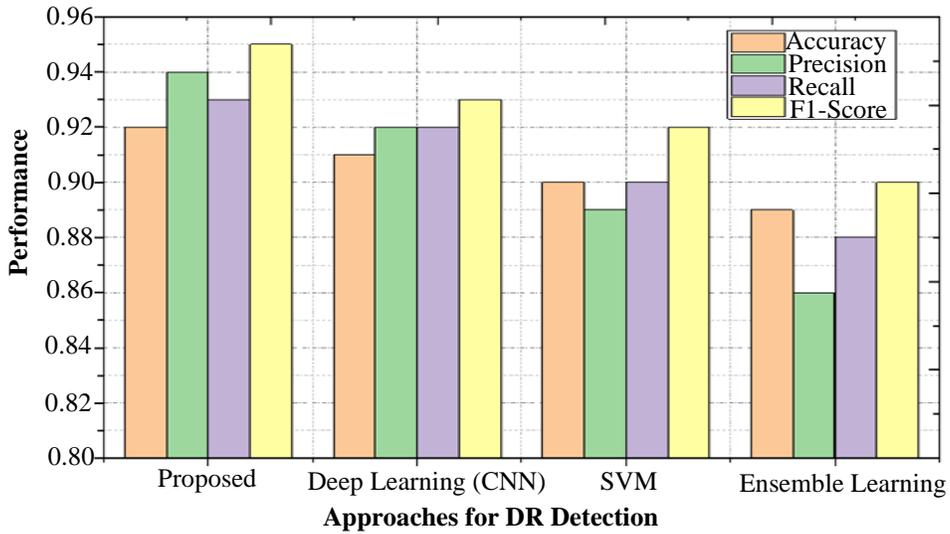

**Fig. 4 Performance comparison - DIARETDB1 database**

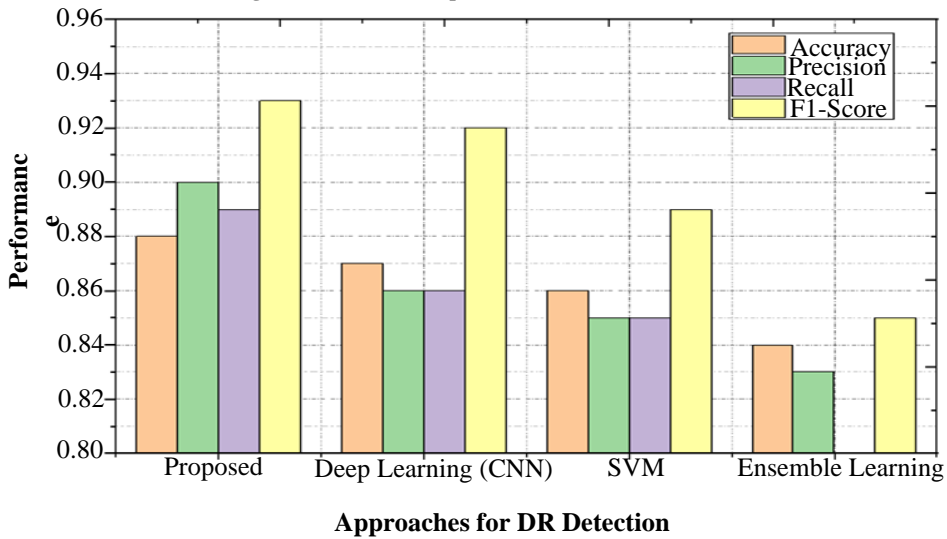

**Fig. 5 Performance comparison - DRISHTI-GS1 database**

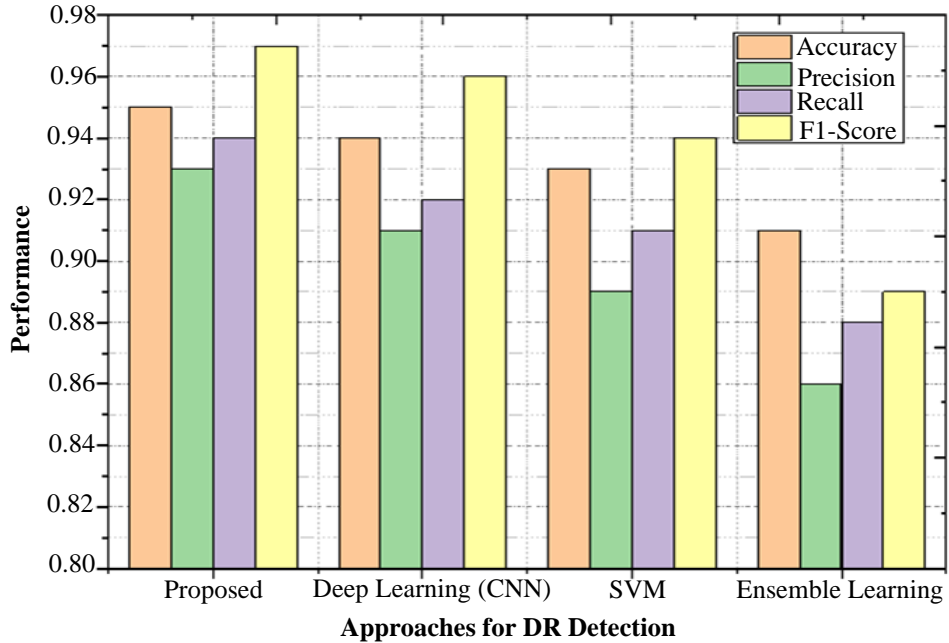

Fig. 6 Performance comparison - e-ophtha ex database

## 5. Conclusion

In conclusion, developing a Semi-Supervised Graph Learning algorithm for Diabetic Retinopathy (DR) detection has exposed auspicious outcomes in improving the accuracy of DR detection. By leveraging the relationships between labelled and unlabelled data in a graph structure, the algorithm can effectively propagate label information, thereby enhancing the accuracy of DR detection. The experimental results of the proposed algorithm have demonstrated that it outperforms several state-of-the-art methods in terms of metrics. The algorithm has also been shown to be robust to noise and outliers, which are common challenges in medical image analysis.

Additionally, the proposed technique can handle imbalanced datasets, which is essential for medical picture analysis because the proportion of positive instances is frequently significantly lower than that of negative ones.